\documentclass{bmvc2k}

\usepackage{graphicx}
\usepackage{amsmath,amssymb} 
\usepackage{multirow}
\usepackage{booktabs}

\usepackage{hyperref}
\usepackage{algorithm}
\usepackage{algcompatible}
\usepackage{soul}
\title{Informed Democracy: Voting-based Novelty Detection for Action Recognition}

\addauthor{Alina Roitberg*}{alina.roitberg@kit.edu}{1}
\addauthor{Ziad Al-Halah*}{ziad.al-halah@kit.edu}{1}
\addauthor{Rainer Stiefelhagen}{rainer.stiefelhagen@kit.edu}{1}

\addinstitution{
 Karlsruhe Institute of Technology,\\
 76131 Karlsruhe,\\
 Germany
}

\runninghead{\scriptsize Roitberg, Al-Halah, Stiefelhagen}{\scriptsize Novelty Detection for Action Recognition}

\def\eg{\emph{e.g}\bmvaOneDot}

\def\etal{\emph{et al}\bmvaOneDot}
\def\ie{\emph{i.e.~}}
\newcommand{\tblref}[1]{\mbox{Table~\ref{#1}}}

\newcommand{\secref}[1]{\mbox{Section~\ref{#1}}}
\let\oldparagraph\paragraph
\renewcommand{\paragraph}[1]{\vspace{-0.4cm} \oldparagraph{#1}}

\renewcommand{\eqref}[1]{\mbox{Eq. \ref{#1}}}
\newif\ifedit

\begin{document}

\maketitle

\begin{abstract}
Novelty detection is crucial for real-life applications. 
While it is common in activity recognition to assume a closed-set setting, \ie test samples are always of training categories, this assumption is impractical in a real-world scenario.
Test samples can be of various categories including those never seen before during training.
Thus, being able to know \emph{what we know} and \emph{what we don't know} is decisive for the model to avoid what can be catastrophic consequences. 
We present in this work a novel approach for identifying samples of activity classes that are not previously seen by the classifier. 
Our model employs a voting-based scheme that leverages the estimated uncertainty of the individual classifiers in their predictions to measure the novelty of a new input sample.
Furthermore, the voting is privileged to a subset of \emph{informed} classifiers that can best estimate whether a sample is novel or not when it is classified to a certain known category.
In a thorough evaluation on UCF-101 and HMDB-51, we show that our model consistently outperforms  state-of-the-art in novelty detection.
Additionally, by combining our model with off-the-shelf zero-shot learning (ZSL) approaches, our model leads to a significant improvement in action classification accuracy for the generalized ZSL setting.
\end{abstract}

\section{Introduction}

Human activity recognition from video is a very active research field, with a long list of potential application domains, ranging from autonomous driving to security surveillance~\cite{aggarwal2011human,poppe2010survey}.
However, the vast majority of published approaches are developed under the assumption that all categories are known a priori \cite{carreira2017quo, simonyan2014two,wang2016temporal, hara2018can,varol2017long, ji20133d}.
This \textit{closed set} constraint represents a significant bottleneck in the real world, where the system will probably encounter samples from various categories including those never seen during development.
The set of possible actions is dynamic by its nature, possibly changing over time.
Hence, collecting and maintaining large scale application-specific datasets of video data is especially costly and impractical.
This raises a crucial need for the developed models to be able to identify cases where they are faced with samples out of their knowledge domain.
In this work, we explore the field of activity recognition under \textit{open set} conditions~\cite{scheirer2014probability,scheirer2013toward,busto2017open}, a setting which has been little-explored before especially in the action recognition domain~\cite{moerland2016knowing}.

In an open world application scenario, an action recognition model should be able to handle three different tasks:
1) the standard classification of previously seen categories;
2) knowledge transfer for generalization to new unseen classes (e.g. through zero-shot learning);
3) and knowing how to automatically discriminate between those two cases.
The third component of an open set model lies in its ability to identify samples from unseen classes (\textit{novelty detection}).
This is closely linked to the classifier's confidence in its own predictions, \ie how can we build models, that know, what they do not know?
A straight-forward way is to employ the \textit{Softmax} output of a neural network (NN) model as the basis for a rejection threshold~\cite{richter2017safe,ramos2017detecting}.
Traditionally, action recognition algorithms focus on maximizing the top-1 performance on a static set of actions.
Such optimization leads to \textit{Softmax} scores of the winning class being strongly biased towards very high values~\cite{nguyen2015deep,szegedy2013intriguing,gal2016dropout,KendallGalCipolla2017Multi}.
While giving excellent results in closed set classification, such overly self-confident models become a burden under open set conditions.
A better way to asses NN's confidence, is to rather predict the probability distribution with Bayesian neural networks (BNN).
Recently, Gal \etal~\cite{gal2016dropout} introduced a way of efficiently approximating BNN modeled as a Gaussian Process~\cite{rasmussen2004gaussian} and using dropout-based Monte-Carlo sampling (MC-Dropout)~\cite{gal2016dropout}.
We leverage the findings of \cite{gal2016dropout} and exploit the \textit{predictive uncertainty} in order to identify activities of previously unseen classes.

This work aims at bringing conventional activity recognition to a setting where new categories might occur at any time and has the following main contributions:
1) We present a new model for novelty detection for action recognition based on the predictive uncertainty of the classifiers.
Our main idea is to estimate the \textit{novelty} of a new sample based on the uncertainty of a selected group of output classifiers in a voting-like manner.
The choice of the voting classifiers depends on how confident they are in relation to the currently predicted class.
2) We adapt zero-shot action recognition models, which are conventionally applied solely on samples of the \textit{unseen} classes, to the generalized case (\ie~open set scenario) where a test sample may originate from either known or novel categories.
We present a generic framework for generalized zero-shot action recognition, where our novelty detection model serves as a filter to distinguish between seen and novel categories, passing the sample either to a standard classifier or a zero-shot model accordingly.
3) We extend the custom evaluation setup for action recognition to the open-set scenario and formalize the evaluation protocol for the tasks of novelty detection and zero-shot action recognition in the generalized case on two well-established datasets, UCF-101~\cite{soomro2012ucf101} and HMDB-51~\cite{kuehne2013hmdb51}.
The evaluation shows, that our model consistently outperforms conventional NNs and other baseline methods in identifying novel activities and was highly successful when applied to generalized zero-shot learning.

\section{Related Work}

\vspace{0.4cm}

\paragraph{Novelty Detection}

Various machine learning methods have been used for quantifying the \textit{normality} of a data sample.
An overview of the existing approaches is provided by \cite{pimentel2014review, chandola2009anomaly}.
A lot of today's novelty detection research is handled from the probabilistic point of view \cite{ liu2017incremental,socher2013zero,pimentel2014review,mohammadi2018conditional},  modeling the probability density function (PDF) of the training data, with Gaussian Mixture Models (GMM) being a popular choice \cite{pimentel2014review}.
The One-class SVM introduced by Sch\"olkopf \etal \cite{scholkopf2001estimating} is another widely used unsupervised method for novelty detection, mapping the training data into the feature space and maximizing the margin of separation from the origin.
Anomaly detection with NNs has been addressed several times using encoder-decoder-like architectures and the reconstruction error~\cite{williams2002comparative}.
A common way for anomaly detection is to threshold the output of the neuron with the highest value \cite{markou2006neural, hendrycks17baseline, richter2017safe}.
Recently, Hendrycks \etal\cite{hendrycks17baseline} presented a baseline for deep-learning based visual recognition using the top-1 Softmax scores and pointed out, that this area is under-researched in computer vision.

The research of novelty detection in videos has been very limited.
A related topic of \textit{anomaly detection} has been studied for very specific applications, such as surveillance \cite{pimentel2014review,markou2006neural} or personal robotics\cite{moerland2016knowing}. 
Surveillance  however often has anomalies, such as \textit{Robbery} or \textit{Vandalism}, present in the training set in some form~\cite{mohammadi2016angry,sultani2018real} which violates our open-set assumption.
The work most similar to ours is the one of Moerland \etal \cite{moerland2016knowing} where Hidden-Markov-Model is used to detect unseen actions from skeleton features.
However, \cite{moerland2016knowing} considers only a simplified evaluation setting using only a single \textit{unseen} action category in testing.
In contrast to  \cite{moerland2016knowing} our model is based on a deep neural architecture for detecting novel actions which makes it applicable to a wide range of modern action recognition models.
Furthermore, we consider a challenging evaluation setting on well-established datasets where novel classes are as diverse as those seen before.
Additionally, we go beyond novelty detection and evaluate how well our model generalizes to classifying novel classes through zero-shot learning.
Our model leverages approximation of BNN using MC-Dropout as proposed by Gal \etal \cite{gal2016dropout},  which has been successfully applied in semantic segmentation~\cite{KendallGalCipolla2017Multi} and active learning~\cite{Gal2017Active}.
We extend the BNN approximation to the context of open set action recognition where we incorporate the uncertainty of the output neurons in a voting scheme for novelty detection.

\paragraph{Zero-Shot Action Recognition}

Research on human activity recognition under open set conditions has been sparse so far.
A related field of Zero-Shot Learning (ZSL) attempts to classify new actions without any training data by linking visual features and the high-level semantic descriptions of a class, \eg through action labels.
The description is often represented with word vectors by a skip-gram model (\eg \textsl{word2vec}~\cite{mikolov2013distributed}) previously trained on a large-scale text corpus.
ZSL for action recognition gained popularity over the past few years and has also been improving slowly but steadily~\cite{xu2017transductive,wang2017zero,xu2015semantic,qin2017zero, zhu2018towards}. 
In all of these works, the categories used for training and testing are disjoint and the method is evaluated on unfamiliar actions only. This is not a realistic scenario, since it requires the knowledge of whether the activity belongs to a known or novel category a priori. Generalized zero-shot learning (GZSL) has been recently studied for image recognition and a drastic performance drop of classical ZSL approaches such as ConSE~\cite{norouzi2013zero} and Devise~\cite{frome2013devise}, has been reported~\cite{xian2017zero}.
As the main application of our novelty detection approach, we implement a framework for ZSL in the generalized case and integrate our novelty detection method to distinguish between known and unknown actions.

\begin{figure}[t]
	\centering
	\includegraphics[width=\textwidth, height = 3.5cm]{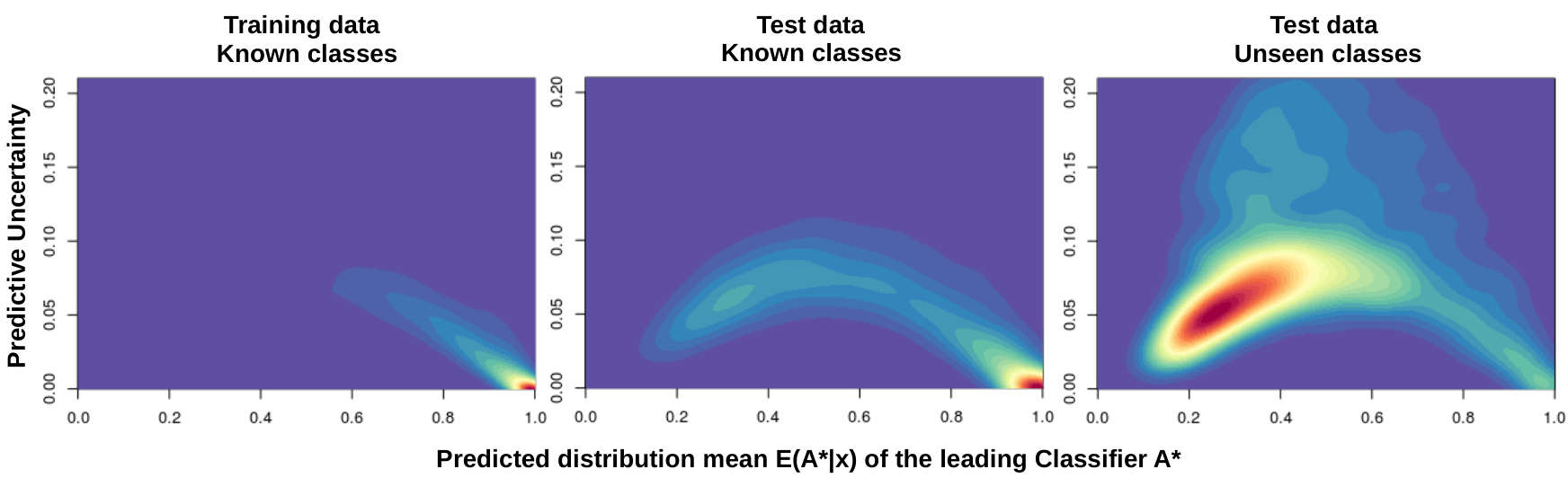}
	\vspace{-0.2cm}
	\caption{Distribution of predictive mean and uncertainty as a 2-D histogram of the leading classifier (highest predictive mean) for the input with known and unseen actions (HMDB-51 dataset). Red denotes common cases (high frequency), blue denotes unlikely cases.  }
	\label{fig:freq_score_comparison}
	\vspace{-0.5cm}
\end{figure}

\section{Novelty Detection via Informed Voting}
\label{sec:approach_novelty}

We present a new approach for novelty detection in action recognition.
That is, given a new video sample $\mathbf{x}$, our goal is to find out whether $\mathbf{x}$ is a sample of a previously known category or if it belongs to a novel action category not seen before during training.

Let $A=\{A_1,...A_K\}$ be the set of all $K$~\emph{known} categories in our dataset.
Then $p(A_i|\mathbf{x})$ is the classifier probability of action category $A_i$ given sample $\mathbf{x}$.
Conceptually, our novelty detection model is composed of two main components: 1) the leader and 2) the council.
The leader refers to the classifier with the highest confidence score in predicting the class of a certain sample $\mathbf{x}$.
For example, in classification neural networks it is common to select the leader based on the highest softmax prediction score.
The leader votes for sample $\mathbf{x}$ being of its own category and assuming that the class of $\mathbf{x}$ is one of the known categories, \ie $\mathrm{class}(\mathbf{x})=A^*\in A$.
The council, on the other hand, is a \textit{subset} of classifiers that will help us validating the decision of a specific leader.
In other words, the council members of a leader representing the selected class $A^*$ are a subset of the classifiers representing the rest of the classes, \ie $C_{A^*}\subseteq A\setminus \{A^*\}$.
These members are elected for each leader individually, \ie each category classifier in our model has its own council.
A council member is selected based on its certainty variance in relation to a leader.
Whenever a leader decides on the category of a sample $\mathbf{x}$, its council will convene and vote on the leader decision.
Then, the council members will jointly decide whether the leader made the correct decision or it was mistaken because the sample is actually from a novel category.

Next, we explain in details how we measure the uncertainty of a classifier (\secref{sec:uncertainty}); choosing a leader and its council members (\secref{sec:council}); and, finally, the novelty voting procedure given new sample (\secref{sec:novelty_vote}).

\subsection{Measuring Classifier Uncertainty}
\label{sec:uncertainty}

In this section, we tackle the problem of quantifying the \textit{uncertainty} of a classifier given a new sample.
The estimated uncertainty is leveraged later by our model to select the council members as we will see in \secref{sec:council}.

In the context of deep learning, it is common to consider the single point estimates for each category, represented by the output of the softmax layer, as a confidence measure~\cite{markou2006neural, hendrycks17baseline, augusteijn2002neural, richter2017safe}.
However, this practice has been highlighted in literature to be inaccurate since a model can be highly uncertain even when producing high prediction scores~\cite{nguyen2015deep,szegedy2013intriguing}.
Bayesian neural networks (BNNs) offer us an alternative to the point estimate models and are known to provide a well calibrated estimation of the network uncertainty in its output.
Given the network parameters $\omega$ and a training set $S$, the predictive probability of the BNN is obtained by integrating over the parameter space.
The prediction $p(A_i|\textbf{x}, S)$ is therefore the mean over all possible parameter combinations weighted by their posterior probability:
\begin{equation}
	p(A_i|\textbf{x}, S) = \int_{\omega}^{}{p(A_i|\textbf{x},\omega)p(\omega|S)d\omega}
\end{equation}
However, BNNs are known to have a difficult inference scheme and high computation cost~\cite{gal2016dropout}.
Therefore, we leverage the robust model proposed by \cite{gal2016dropout} to approximate the predictive mean and uncertainty of the BNN posterior distribution with network parameters modeled as a Gaussian Process (GP). 
This method is based on dropout regularization~\cite{srivastava2014dropout}, a widely used technique which has proven to be very effective against overfitting.
That is, it leverages the dropout at each layer in the network to draw the weights from a Bernoulli distribution with probability $p$.
At test time, the dropout is iteratively applied for $M$ forward passes for each individual sample.
Then, the statistics of the neural network output represents a Monte-Carlo (MC) approximation of the neuron's posterior distribution
This approach is referred to as MC-Dropout~\cite{gal2016dropout}.

Specifically, let $\mathbf{x}$ be a representation generated by a convolutional neural network (CNN) for an input sample $z$.
We add a feedforward network on top of the CNN with two fully-connected layers with weight matrices $W_1$ and $W_2$.
Instead of using a deterministic \textit{Softmax} estimate in a single forward pass as it is common with CNNs, we now compute the mean over $M$ stochastic iterations as our prediction score:
\begin{equation}\label{eq:mean}
\mathbb{E}(A_i|\textbf{x}) \approx  \frac{1}{M}\sum_{m=1}^{M}  \mathrm{softmax}(\mathrm{relu}(\mathbf{x}^T D_1 W_1+\mathbf{b}_1)D_2W_2),
\end{equation}
where $relu(\cdot)$ is the rectified linear unit (ReLU) activation function, $\mathbf{b}_1$ is the bias vector of the firs layer.
Additionally, $D_1$ and $D_2$ are diagonal matrices where the diagonal elements contain binary values, such that they are set to $1$ with probability $1-p$ and otherwise to $0$.

We further empirically compute the model's \textit{predictive uncertainty} as the distribution variance:
\begin{equation}
	U(A_i|\textbf{x}) \approx s^2 = \frac{1}{M-1}\sum_{m=1}^{M} [\mathrm{softmax}(\mathrm{relu}(\mathbf{x}^TD_1W_1+\mathbf{b}_1)D_2W_2) -\mathbb{E}(A_i|\textbf{x}) ]^2
\end{equation}

Fig. \ref{fig:freq_score_comparison} shows how predictive mean and uncertainty are distributed for samples of known and novel classes.
The plot depicts clearly different patterns for the resulting probability distributions in these two cases which illustrates the potential of Bayesian uncertainty for novelty detection.

\subsection{Selecting the Leader and its Council}
\label{sec:council}

\begin{figure}[t]
	\centering
	\includegraphics[width=0.99\textwidth]{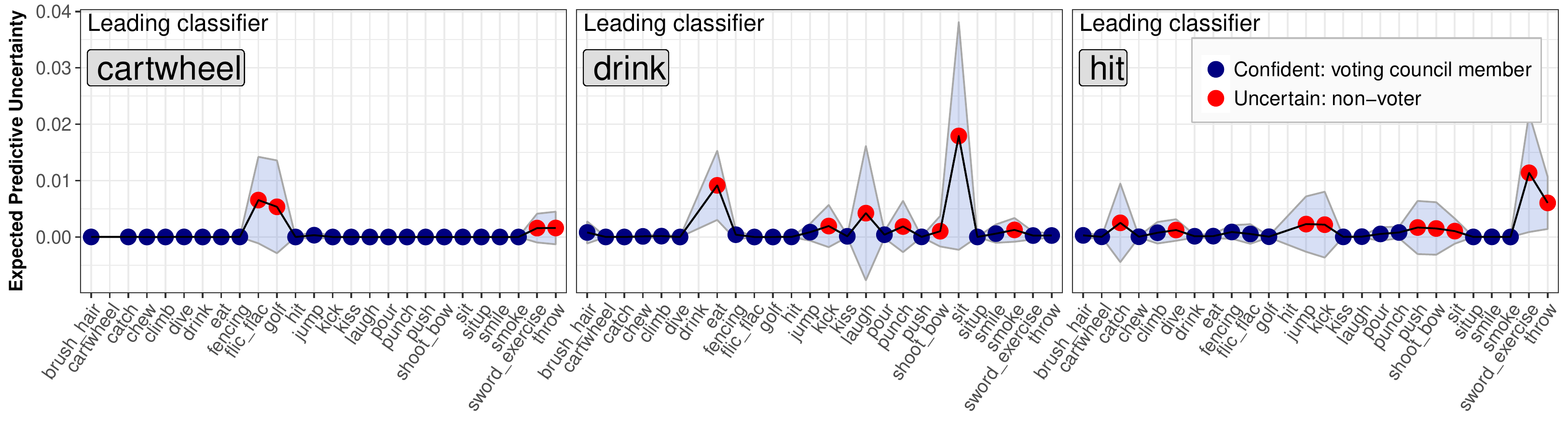}
	\vspace{-0.2cm}
    \caption{Council members and uncertainty statistics for three different leaders (HMDB-51).
    The classifier's \emph{average} uncertainty  and its \emph{variance} (area surrounding the point) illustrate how it changes its belief in the leader for different data inputs.
    Blue points are in the council of the current leader, while red points are classifiers that did not pass the credibility threshold.  }
	\label{fig:anchor_points}
    \vspace{-0.5cm}
\end{figure}

Now that we can estimate the confidence and uncertainty of each category classifier in our model, we describe in this section how to choose the leader and select it council members.

\paragraph{The Leader.}
Rather than selecting the leader using a point estimate based on the softmax scores of the output layer, we leverage here the more stable dropout-based estimation of the prediction mean.
Hence, the leader is selected as the classifier with highest expected prediction score over $M$ sampling iterations:
\begin{equation}
	A^* = \underset{A_k \in A}{\operatorname{argmax}} \ {\mathbb{E}(A_i|\textbf{x})},
\end{equation}
where $\mathbb{E}(A_i|\textbf{x})$ is estimated according to \eqref{eq:mean}.

\paragraph{The Council.}
The leader by itself can sometimes produce highly confident predictions for samples of unseen categories~\cite{szegedy2013intriguing}.
Hence, we can not rely solely on the leader confidence to estimate whether a sample is of a novel category or not.
Here, the rest of the classifiers can help in checking the validity of the leader's decision.
We notice that these classifier exhibit unique patterns in regard to a certain leader.
They can be grouped into two main groups: the first shows high uncertainty when the leader is correctly classifying a sample;
while the second shows a very low uncertainty and are in agreement with the leader.

Guided by this observation, we select the members of the Council $C_A^*$ for a certain leader $A^*$ based on their uncertainty variance in regards to samples of the leader's category, \ie $\mathbf{x}\in A^*$.
In other words, those classifiers that exhibit very low uncertainty when the leader is classifying samples of its own category are elected to join its council.
During the training phase, we can select the council members for each classifier in our model.
Here, we randomly split the initial set into a training set $S_{train}$ which is used for model optimization and parameter estimation,
and a holdout set $S_{holdout}$ which is used for choosing the council member for all the classifiers iteratively.
Specifically, we use a 9/1 split for the training and the holdout splits.
We first estimate the parameters of our deep model $\omega$ using $S_{train}$. 
Then, we evaluate our model over all samples from $S_{holdout}$.
For each category classifier in our model, we construct a set of true positive samples $S_{tp}^{A_i}\subseteq S_{holdout}$.
For each sample $\mathbf{x_n}\in S_{tp}^{A_i}$, we estimate the uncertainty $U(A_j|\mathbf{x_n})$ of the rest of the classifiers $A_j\in A\setminus \{A_i\}$ using the MC-Dropout approach.
Then, the variance of these classifiers' uncertainty is estimated as:
\begin{equation}
	Var(A_j|A_i) = \frac{1}{N}\sum_{n=1}^{N} (U(A_j|\mathbf{x}_n) -\mathbb{E}[U(A_j|\textbf{x})] )^2
\end{equation}
where $N=|S_{tp}^{A_i}|$ and $\mathbb{E}[U(A_j|\textbf{x})]$ is the expectation of the uncertainty of the classifier $A_j$ over samples $\mathbf{x}\in S_{tp}^{A_i}$.
Finally, classifiers with a variance lower than a fixed credibility threshold $Var(A_j|A_i) < c$ are then elected as members of $A_i$ council.

Fig. \ref{fig:anchor_points} shows three leaders and their elected councils according to our approach.
We see, for example, that eight classifiers did not pass the credibility threshold for the leader \textsl{drink} and were excluded from its council.
The variance of the uncertainty is especially high for \textsl{sit} and \textsl{eat} in this case.
This is expected since those actions often occur in a similar context.

\subsection{Voting for Novelty}
\label{sec:novelty_vote}

Given the trained deep model and the sets of all council members from the previous step, we can now generate a novelty score for a new sample $\mathbf{x}$ as follows.
First, we calculate the prediction mean $\mathbb{E}[p(A_i|\mathbf{x})]$ and uncertainty $U(A_i|\mathbf{x})$ of all the action classifiers using $M$ stochastic forward passes and MC-Dropout.
Then, the classifier with the maximum predicted mean is chosen as the leader.
Finally, the council members of the chosen leader vote for the novelty of sample $\mathbf{x}$ based on their estimated uncertainty (see Algorithm~\ref{alg:voting}).

\begin{figure}
	\centering     
	\includegraphics[width=0.32\textwidth]{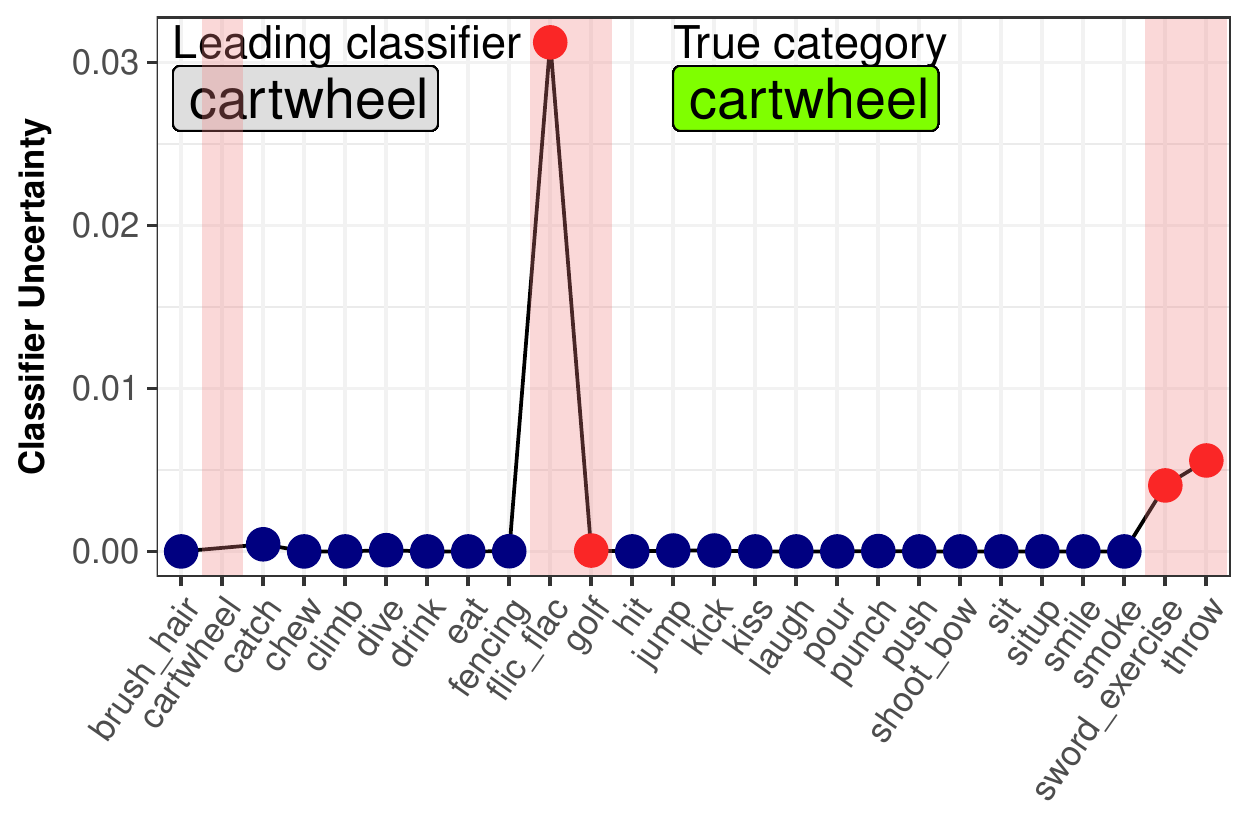}
	\includegraphics[width=0.32\textwidth]{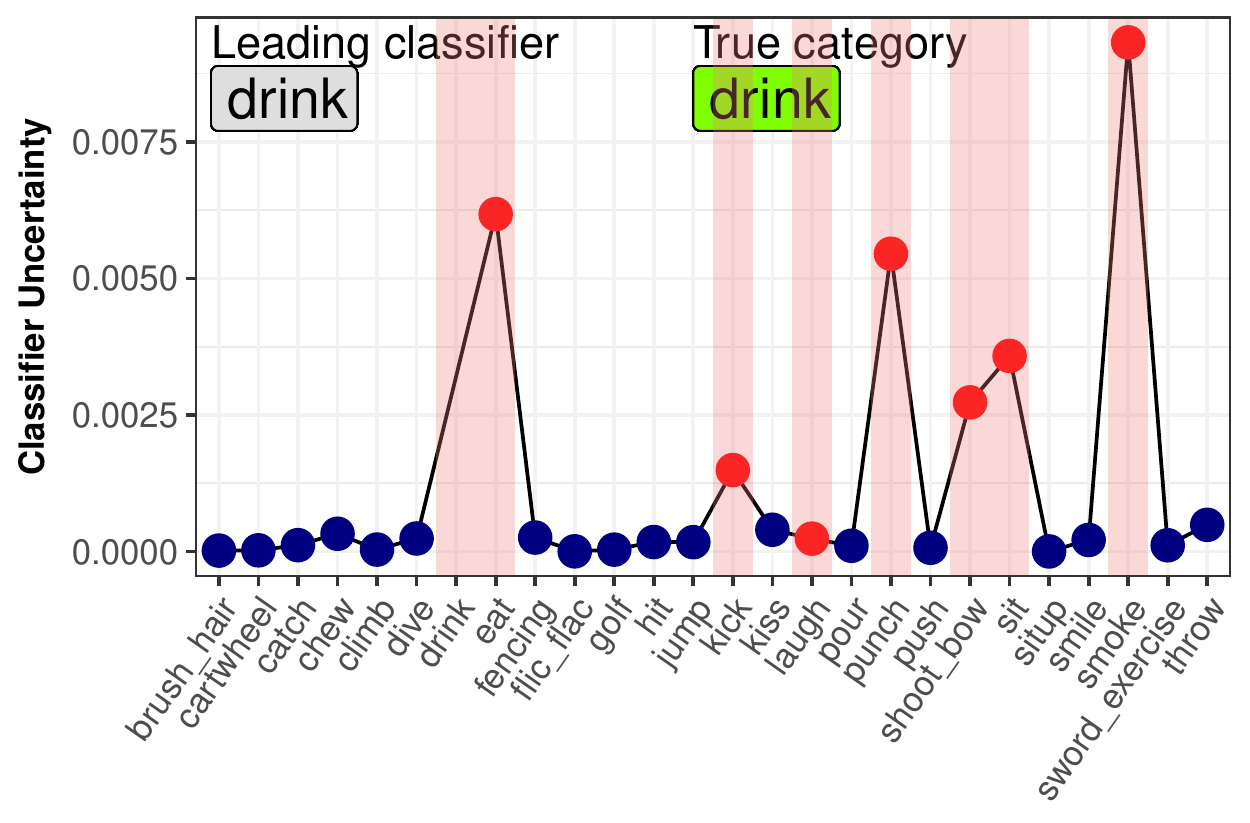}
	\includegraphics[width=0.32\textwidth]{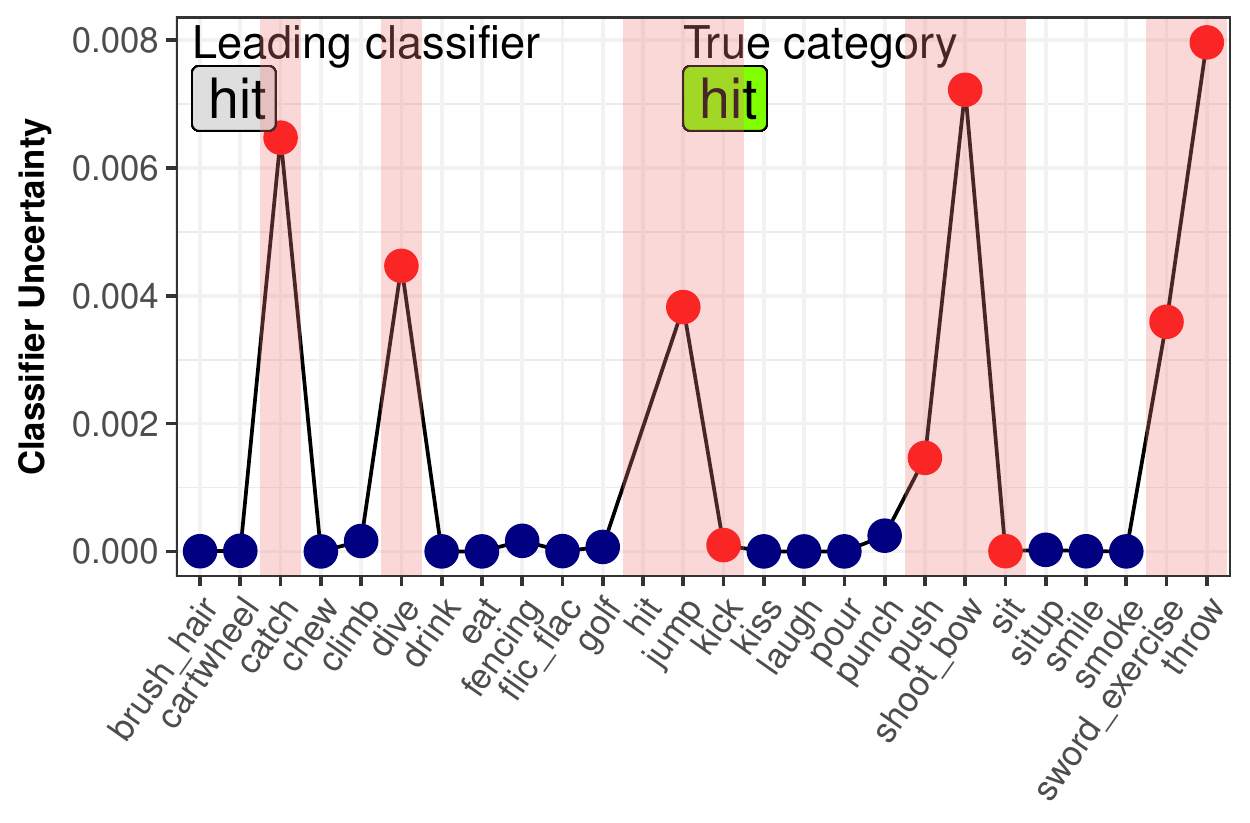}

	\includegraphics[width=0.32\textwidth]{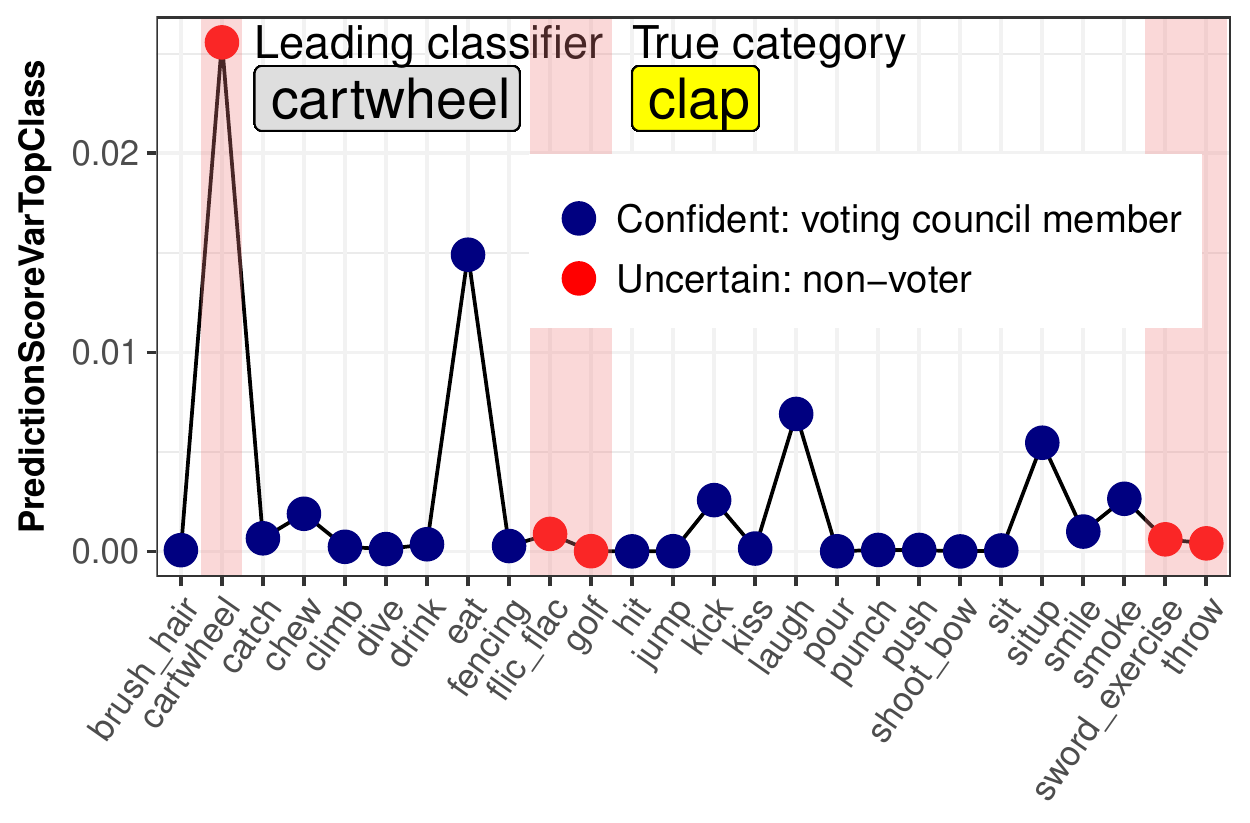}
	\includegraphics[width=0.32\textwidth]{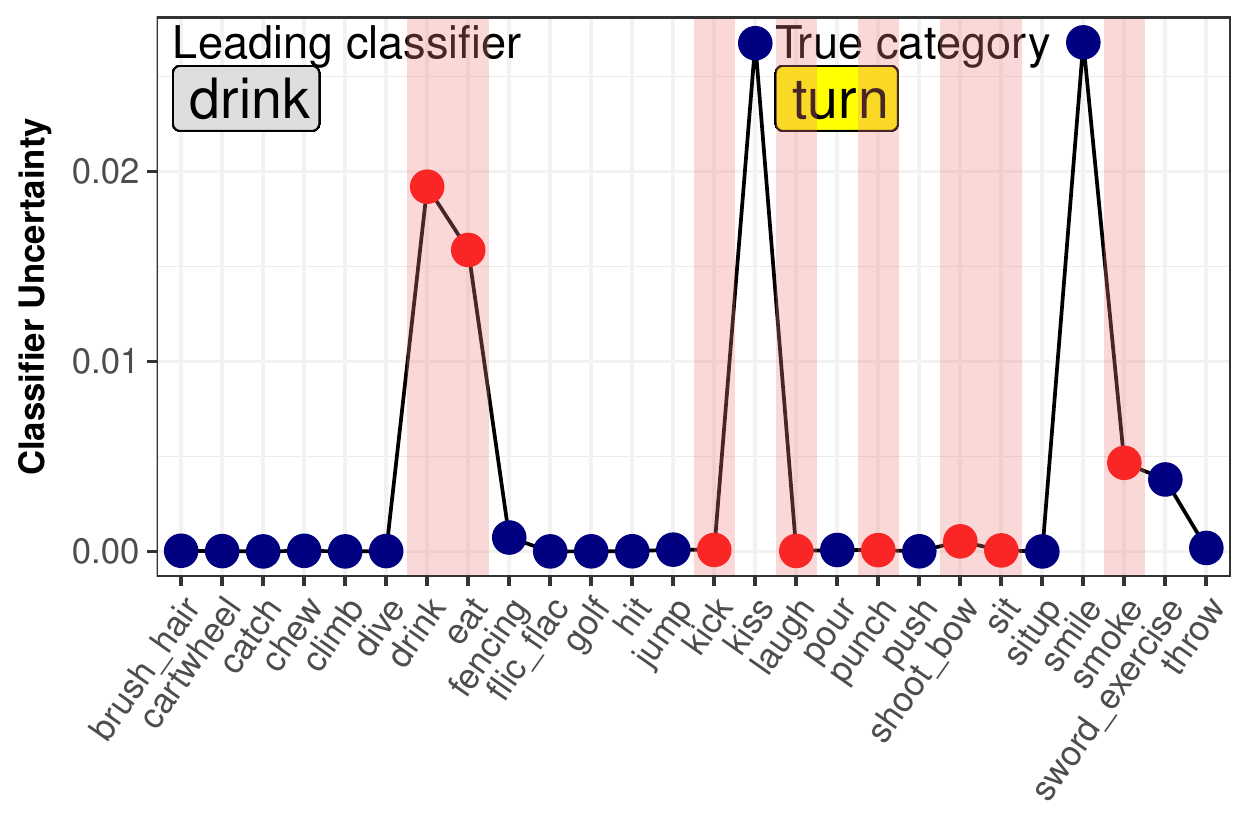}
	\includegraphics[width=0.32\textwidth]{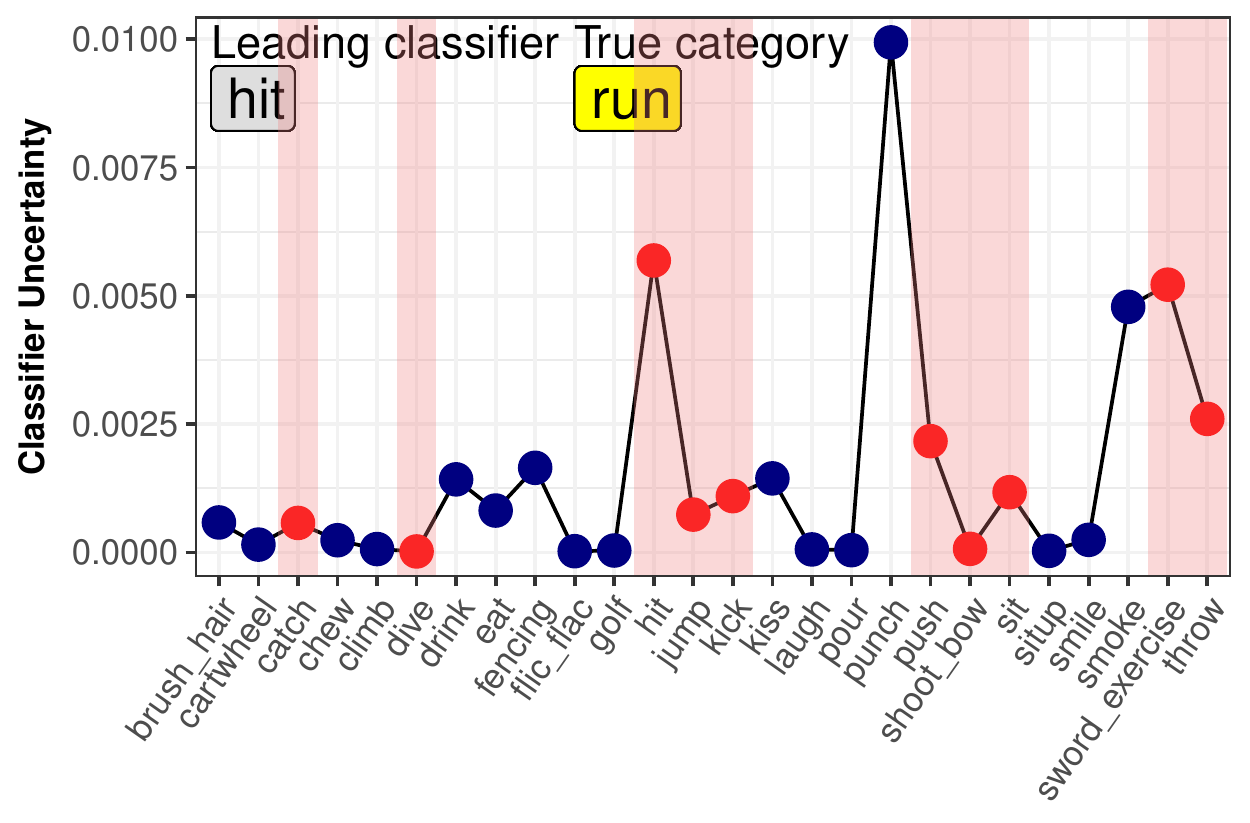}
	\vspace{-0.2cm}
	\caption{Examples of selective voting for the novelty score of different activities.
	The first row depicts the case where the samples are of \emph{known} classes and second row for those of \emph{novel} classes.
	Red points highlight classifiers, which were \emph{excluded} from to the council of the current leader. Their uncertainty is, therefore, ignored when inferring the novelty score.  }
	\label{fig:examples_seen_unseen}
	\vspace{-0.5cm}
\end{figure}

Examples of such voting outcome for three different leaders are illustrated in Fig. \ref{fig:examples_seen_unseen}.
In case of category \textsl{cartwheel}, we can see that when the leader is voting indeed for the correct category, all council members show low uncertainty values therefore resulting in a low novelty score, as uninformed classifiers (marked in red) are excluded.
However, we observe very different measurements for an example from an unseen category \textsl{clap} which is also predicted as \textsl{cartwheel}.
Here, multiple classifiers which are in the council (marked in blue) show unexpected high uncertainty values (e.g. \textsl{eat, laugh}), therefore discrediting the leader decision and voting for a high novelty score.

\begin{algorithm}[]
	\label{alg:voting}
	\renewcommand{\thealgorithm}{}
	\caption{Novelty Detection by Voting of the Council Neurons}

	\begin{algorithmic}[1]
		\STATEx \textbf{Input:} Input sample $\textbf{x}$, Classification Model $\omega$, $K$ sets of \textsl{Council} members for each \textsl{Leader}: $\{C_{A^1},...,C_{A^K}\}$
		\STATEx \textbf{Output:} Novelty score  $\upsilon(\vec{x})$
		\STATE \textbf{Inference using MC-Dropout}

		Preform $M$ stochastic forward passes: $p_{A_i}^{m} = p(A_i|\textbf{x}, \omega^m)$;
		\FORALL{$A_i \in A$}
		\STATE{Calculate the prediction mean and uncertainty: $\mathbb{E}(p(A_i|\textbf{x}))$ and  $U(A_i|\textbf{x})$} \ENDFOR
		\STATE {	Find the \textsl{Leader}:\quad	$A^* = \underset{A_k \in A}{\operatorname{argmax}} \ {p(A_i|\textbf{x})}$
		}
		\STATE{ Select the \textsl{Council}:\quad $C_{A*}$}
		\STATE Compute the \textit{novelty score}  :
	$\upsilon(\textbf{x}) = \frac{\sum_{A_i \in C_{A^*}}{U(A_i|\textbf{x})}} {|C_{A^*}|}$

	\end{algorithmic}
\end{algorithm}

\paragraph{Model variants.}
We refer to our previous model as the \emph{Informed Democracy} model since voting is restricted to the council members which are chosen in an informed manner to check the decision of the leader.
In addition to the previous model, we consider two other variants of our model:
\begin{enumerate}
\item The \emph{Uninformed Democracy} model: Here, there is no council and all classifiers have the right to vote for any leader.
Hence, step $7$ in Algorithm~\ref{alg:voting} is replaced with 	$\upsilon(\textbf{x}) = \frac{\sum_{A_i \in A}{U(A_i|\textbf{x})}} {K}$.
\item The \emph{Dictator} model: unlike the previous model, this one leverages only the leader's uncertainty in its own decision to predict the novelty of the sample, \ie $\upsilon = U(A^*|\vec{x})$.
\end{enumerate}

\paragraph{Open set and zero-shot learning}
Once our model generated the novelty score $\upsilon(\mathbf{x})$, we can decide whether $\mathbf{x}$ is a sample from a novel category or not using a sensitivity threshold $\tau$.
This threshold can be estimated from a validation set using the equal error rate of the receiver operating characteristic curve (ROC).
Then, if $\upsilon(\mathbf{x}) < \tau$  the \textsl{Council} votes in favor of the \textsl{Leader} and its category is taken as our final classification result.
Otherwise, an \textit{unknown} activity class has been identified.
In this case, the input could be passed further to a module in charge of handling unfamiliar data, such as a zero-shot learning model or a user to give the sample a new label in the context of active learning.

\section{Evaluation}

\vspace{0.4cm}

\paragraph{Evaluation setup }

Since there is no established evaluation procedure available for action recognition in open-set conditions, we adapt existing evaluation protocols for two well-established datasets, HMDB-51~\cite{kuehne2013hmdb51} and UCF-101~\cite{soomro2012ucf101}, for our task\footnote[2]{Dataset splits used for novelty detection and generalized zero-shot action recognition are provided  at \href{https://cvhci.anthropomatik.kit.edu/~aroitberg/novelty_detection_action_recognition}{https://cvhci.anthropomatik.kit.edu/$\sim$aroitberg/novelty\_detection\_action\_recognition}}.
We evenly split each dataset into seen/unseen categories (26/25 for HMDB-51 and 51/50 for UCF-101).
Samples of unseen classes will not be available during training, while samples of the remaining set of seen classes is further split into training (70\%) and testing (30\%) sets, thereby adapting the evaluation framework of \cite{wang2017zero} for the \textit{generalized} ZS learning scenario.
For each dataset, we randomly generate 10 splits and report the average and standard deviation of the recognition accuracy.
Using a separate validation split, we optimize the credibility threshold $c$ and compute the threshold for rejection $\tau$ for each category as the Equal Error Rate of the ROC.

\paragraph{Architecture details}

We augment the RGB-stream of the I3D architecture \cite{carreira2017quo} with MC-Dropout.
The model is pre-trained on the Kinetics dataset, as described in \cite{carreira2017quo}.
The last average pooling is connected to two fully connected layers: a hidden layer of size 256 and the final softmax-classifier layer.
These are optimized using SGD with momentum of 0.9, learning rate of 0.005 and dropout probability of 0.7 for 100 epochs.
We sample the output scores for $M=100$ stochastic forward passes applied on the two layers preceding the classifier, while the credibility threshold $c$ is set to 0.001.

\paragraph{Baselines}

We compare our model to three  popular methods for novelty and outlier detection:
1) a One Class SVM \cite{scholkopf2000support,scholkopf2001estimating} with RBF kernel (upper bound on the fraction of training errors $\nu$ set to 0.1);
2) a GMM \cite{zorriassatine2005novelty, pimentel2014review} with 8 components;
3) and Softmax probabilities \cite{richter2017safe, hendrycks17baseline} as the value for thresholding.
Both SVM and GMM were trained on normalized features obtained from last average pooling layer of I3D pre-trained on the Kinetics dataset~\cite{carreira2017quo}.

\paragraph{Novelty Detection}

We evaluate the novelty detection accuracy in terms of a binary classification problem, using the area under curve (AUC) values of the receiver operating characteristic (ROC) and the precision-recall (PR) curves.

We show the robustness of our approach in comparison to the baseline methods in \tblref{tbl:novelty}.
All variants of our model clearly outperform the conventional approaches and achieve an ROC-AUC gain of over 7\% on both datasets. 
Along our model variants, \textit{Informed Democracy} has proven to be the most effective strategy for novelty score voting, outperforming the \textit{Dictator} by 5.5\% and 1.4\%, while \textit{Uninformed Democracy} achieved second-best results.
We believe that smaller differences in performance gain on the UCF-101 data are due to the much higher supervised classification accuracy on this dataset.
Since the categories of UCF-101 are easier to distinguish visually and the confusion is low, there is more agreement between the neurons in terms of their confidence.

\paragraph{Generalized Zero-Shot Learning (GZSL)}

Next, we evaluate our approach in the context of GZSL, where our novelty detection model serves as a filter to distinguish whether the observed example should be classified with the I3D model in the standard classification setup, or mapped to one of the unknown classes via a ZSL model.
We compare two prominent ZSL methods: ConSE \cite{norouzi2013zero} and DeViSE \cite{frome2013devise}.
The ConSE model starts by predicting probabilities of the seen classes, and then takes the convex combination of word embeddings of the top K most possible seen classes and select its nearest neighbor from the novel classes in the \textsl{word2vec} space.
For DeViSE, we train a separate model to regress \textsl{word2vec} representations from the visual features.
We use the publicly available \textsl{word2vec} model that is trained on Google News articles~\cite{mikolov2013distributed}.

For consistency, we first report the results for the standard ZS case (\ie U$\rightarrow$U) and further extend to the generalized case (\ie U$\rightarrow$U+S and U+S$\rightarrow$U+S) as shown in \tblref{tbl:zs}.
In the more realistic GZSL setup, our model is not restricted to any group of target labels and is evaluated on actions of seen and unseen category using the \textit{harmonic mean} of accuracies for seen and unseen classes as proposed by~\cite{xian2017zero}.
\tblref{tbl:zs} shows a clear advantage of employing novelty detection as part of a GZSL framework.
While failure of the original ConSE and DeViSE models might be surprising at first glance, such performance drops have been discussed in previous work on ZSL for image recognition \cite{xian2017zero} and is due to the fact that both models are biased towards labels that were used during training.
Our \textit{Informed Democracy} model yields the best recognition rates in every setting and can therefore be indeed successfully applied for multi-label action classification in case of new activities.

\begin{table}[]
	\centering
	\resizebox{0.9\textwidth}{!}{
		\begin{tabular}{ l l l l l }
			\toprule
			\multirow{2}{*}{\textbf{Novelty Detection Model}} & \multicolumn{2}{c}{HMDB-51}                    & \multicolumn{2}{c}{UCF-101}                    \\
															  & ROC AUC \%             & PR AUC \%              & ROC AUC \%             & PR AUC \%              \\
			\midrule
			\textbf{Baseline Models}                          &                        &                        &                        &                        \\
			One-class SVM                                     & 54.09 ($\pm$3.0)          & 77.86 ($\pm$4.0)    & 53.55 ($\pm$2.0)       & 78.57 ($\pm$2.4)          \\
			Gaussian Mixture Model                            & 56.83 ($\pm$4.2)          & 78.40 ($\pm$3.6)    & 59.21 ($\pm$4.2)       & 79.50 ($\pm$2.2)           \\
			Conventional NN Confidence                        & 67.58 ($\pm$3.3)          & 84.21 ($\pm$3.0)    & 84.28 ($\pm$1.9)       & 93.92 ($\pm$0.7)          \\
			\midrule
			\multicolumn{3}{l}{\textbf{Our Proposed Model based on Bayesian Uncertainty}}                         &                        &                        \\
			Dictator					                      & 71.78 ($\pm$1.8)          & 86.81 ($\pm$2.5)    & 91.43 ($\pm$2.3)       & 96.72 ($\pm$1.0)          \\
			Uninformed Democracy                              & 73.81 ($\pm$1.7)          & 87.83 ($\pm$2.3)    & 92.13 ($\pm$1.8)       & 97.15 ($\pm$0.7)          \\
			Informed Democracy                                & \textbf{75.33 ($\pm$2.7)} & \textbf{88.66 ($\pm$2.3)} & \textbf{92.94 ($\pm$1.7)} & \textbf{97.52 ($\pm$0.6)} \\
			\bottomrule
		\end{tabular}
	}
	\caption{Novelty detection results evaluated as area under the ROC and PR-curves for identifying previously unseen categories (mean and standard deviation over ten dataset splits).}
	\label{tbl:novelty}
\end{table}
 
\begin{table}[]
	\centering

	\resizebox{\textwidth}{!}{
\begin{tabular}{l lll lll }
	\toprule
	\multicolumn{1}{c}{\multirow{2}{*}{\textbf{Zero-Shot Approach}}} & \multicolumn{3}{c}{HMDB-51}                                                                                                     & \multicolumn{3}{c}{UCF-101}                                                                                                     \\
	\multicolumn{1}{c}{}                                             & \multicolumn{1}{c}{U$\rightarrow$U} & \multicolumn{1}{c}{U$\rightarrow$U+S}     & \multicolumn{1}{c}{U+S$\rightarrow$U+S} & \multicolumn{1}{c}{U$\rightarrow$U} & \multicolumn{1}{c}{U$\rightarrow$U+S}     & \multicolumn{1}{c}{U+S$\rightarrow$U+S} \\
	\midrule
	Standard ConSe Model                                               & \multicolumn{1}{l}{21.03 ($\pm$2.07)}    & \multicolumn{1}{l}{0 ($\pm$0)}                 & 0 ($\pm$0)                                     & \multicolumn{1}{l}{17.85 ($\pm$1.95)}    & \multicolumn{1}{l}{0.07 ($\pm$0.10)}            & 0.13 ($\pm$0.20)                                \\
	Standard Devise Model                                              & \multicolumn{1}{l}{17.27 ($\pm$2.01)}    & \multicolumn{1}{l}{0.26 ($\pm$0.37)}           & 0.52 ($\pm$0.73)                               & \multicolumn{1}{l}{14.48 ($\pm$1.13)}    & \multicolumn{1}{l}{0.81 ($\pm$0.36)}           & 1.61 ($\pm$0.71)                               \\
	\midrule
	\textbf{ConSe + Novelty Detection}                                 & \multicolumn{1}{l}{}                 	  & \multicolumn{1}{l}{}                       &                                            & \multicolumn{1}{l}{}                 & \multicolumn{1}{l}{}                       &                                            \\
	One-class SVM                                                      & \multicolumn{1}{l}{21.03 ($\pm$2.07)}    & \multicolumn{1}{l}{10.99 ($\pm$1.83)}          & 17.40 ($\pm$2.41)                               & \multicolumn{1}{l}{17.85 ($\pm$1.95)}    & \multicolumn{1}{l}{10.37 ($\pm$1.59)}          & 16.55 ($\pm$1.91)                              \\
	Gaussian Mixture Model                                             & \multicolumn{1}{l}{21.03 ($\pm$2.07)}    & \multicolumn{1}{l}{13.30 ($\pm$2.58)}           & 19.91 ($\pm$3.32)                              & \multicolumn{1}{l}{17.85 ($\pm$1.95)}    & \multicolumn{1}{l}{9.31 ($\pm$1.30)}            & 15.98 ($\pm$1.99)                              \\
	Conventional NN Confidence                                         & \multicolumn{1}{l}{21.03 ($\pm$2.07)}    & \multicolumn{1}{l}{10.96 ($\pm$0.87)}          & 18.56 ($\pm$1.22)                              & \multicolumn{1}{l}{17.85 ($\pm$1.95)}    & \multicolumn{1}{l}{12.19 ($\pm$1.72)}          & 20.91 ($\pm$2.59)                              \\
	Informed Democracy (ours)                                               & \multicolumn{1}{l}{21.03 ($\pm$2.07)}    & \multicolumn{1}{l}{\textbf{13.67 ($\pm$1.31)}} & \textbf{22.27 ($\pm$1.79)}                     & \multicolumn{1}{l}{17.85 ($\pm$1.95)}    & \multicolumn{1}{l}{\textbf{13.62 ($\pm$1.94)}} & \textbf{23.42 ($\pm$2.97)}                     \\
	\midrule
	\textbf{Devise + Novelty Detection}                                & \multicolumn{1}{l}{}                 	  & \multicolumn{1}{l}{}                       &                                            & \multicolumn{1}{l}{}                 & \multicolumn{1}{l}{}                       &                                            \\
	One-class SVM                                                      & \multicolumn{1}{l}{17.27 ($\pm$2.01)}    & \multicolumn{1}{l}{8.92 ($\pm$1.89)}           & 14.67 ($\pm$2.74)                              & \multicolumn{1}{l}{14.48 ($\pm$1.13)}    & \multicolumn{1}{l}{8.65 ($\pm$1.59)}           & 14.25 ($\pm$2.00)                                 \\
	Gaussian Mixture Model                                             & \multicolumn{1}{l}{17.27 ($\pm$2.01)}    & \multicolumn{1}{l}{10.61 ($\pm$2.22)}          & 16.72 ($\pm$3.1)                               & \multicolumn{1}{l}{14.48 ($\pm$1.13)}    & \multicolumn{1}{l}{7.26 ($\pm$0.84)}           & 12.88 ($\pm$1.40)                               \\
	Conventional NN Confidence                                         & \multicolumn{1}{l}{17.27 ($\pm$2.01)}    & \multicolumn{1}{l}{8.68 ($\pm$1)}              & 15.17 ($\pm$1.56)                              & \multicolumn{1}{l}{14.48 ($\pm$1.13)}    & \multicolumn{1}{l}{10.08 ($\pm$1.59)}          & 17.69 ($\pm$2.33)                              \\
	Informed Democracy (ours)                                               & \multicolumn{1}{l}{17.27 ($\pm$2.01)}    & \multicolumn{1}{l}{\textbf{10.73 ($\pm$1.47)}} & \textbf{18.18 ($\pm$2.21)}                     & \multicolumn{1}{l}{14.48 ($\pm$1.13)}    & \multicolumn{1}{l}{\textbf{11.03 ($\pm$1.42)}} & \textbf{19.48 ($\pm$2.21)}                     \\
	\bottomrule
\end{tabular}
	}
	\caption{ Accuracy for GZS action recognition with the proposed novelty detection model.
	U$\rightarrow$U: test set consists of unseen actions, the prediction labels are restricted to the unseen labels (standard).
	U$\rightarrow$U+S: test set consists of unseen actions, both unseen and seen labels are possible for prediction.
	U+S$\rightarrow$U+S: generalized ZSL case, both unseen and seen categories are among the test examples and in the set of possible prediction labels (harmonic mean of the seen and unseen accuracies reported.)}
	\label{tbl:zs}
\end{table}

\section{Conclusion}
We introduce a new approach for novelty detection in action recognition.
Our model leverages the estimated uncertainty of the category classifiers to detect samples from novel categories not encountered during training.
This is achieved by selecting a council of classifiers for each leader (\ie the most confident classifier).
The council will validate the decision made by the leader through voting.
Hence, either confirming the classification decision for a sample of a known category or revoking the leader decision and deeming the sample to be novel.
We show in a thorough evaluation on two challenging benchmark, that our model outperforms the state-of-the-art in novelty detection.
Furthermore, we demonstrate that our model can be easily integrated in a generalized zero-shot learning framework.
Combining our model with off-the-shelf zero-shot approaches leads to significant improvements in classification accuracy.

\paragraph{Acknowledgements}

This work has been partially funded by the German Federal Ministry of Education and Research (BMBF) within the PAKoS project.

\bibliography{egbib}
\end{document}